\documentclass[letterpaper]{article} 
\usepackage{aaai25}  
\usepackage{times}  
\usepackage{helvet}  
\usepackage{courier}  
\usepackage[hyphens]{url}  
\usepackage{graphicx} 
\usepackage{natbib}  
\usepackage{caption} 
\usepackage{algorithm}
\usepackage{algorithmic}
\usepackage{enumitem}
\usepackage{amsfonts}
\usepackage{amsmath}
\usepackage{amsthm}
\usepackage{multirow}
\usepackage[table,xcdraw]{xcolor}
\newtheorem*{remark}{Remark}
\theoremstyle{definition}
\newtheorem{definition}{Definition}[]

\usepackage{newfloat}
\usepackage{listings}
\DeclareCaptionStyle{ruled}{labelfont=normalfont,labelsep=colon,strut=off} 
\floatstyle{ruled}
\newfloat{listing}{tb}{lst}{}
\floatname{listing}{Listing}
\title{Towards Effective, Efficient and Unsupervised Social Event Detection in the Hyperbolic Space}
\author{
    Xiaoyan Yu\textsuperscript{\rm 1},
    Yifan Wei\textsuperscript{\rm 2},
    Shuaishuai Zhou\textsuperscript{\rm 3}, 
    Zhiwei Yang\textsuperscript{\rm 4}, 
    Li Sun\textsuperscript{\rm 5}, 
    Hao Peng\textsuperscript{\rm 2}, \\
    Liehuang Zhu\textsuperscript{\rm 1}\thanks{Corresponding author.}, 
    Philip S. Yu\textsuperscript{\rm 6}
}
\affiliations{
    \textsuperscript{\rm 1}School of Computer Science and Technology, Beijing Institute of Technology, Beijing, China\\
    \textsuperscript{\rm 2}Beihang University, Beijing, China 
    \enspace
    \textsuperscript{\rm 3}Kunming University of Science and Technology, Kunming, China\\
    \textsuperscript{\rm 4}Institute of Information Engineering, Chinese Academy of Sciences, Beijing, China \\
    \textsuperscript{\rm 5}North China Electric Power University, Beijing, China
    \enspace
    \textsuperscript{\rm 6}University of Illinois at Chicago, Chicago, USA\\
    
    \{xiaoyan.yu, liehuangz\}@bit.edu.cn, \{weiyifan, penghao\}@buaa.edu.cn, psyu@uic.edu
}

\begin{document}

\maketitle

\begin{abstract}

The vast, complex, and dynamic nature of social message data has posed challenges to social event detection (SED). 
Despite considerable effort, these challenges persist, often resulting in inadequately expressive message representations (ineffective) and prolonged learning durations (inefficient). 
In response to the challenges, this work introduces an unsupervised framework, \textbf{HyperSED} (\textbf{Hyper}bolic \textbf{SED}). 
Specifically, the proposed framework first models social messages into semantic-based message anchors, and then leverages the structure of the anchor graph and the expressiveness of the hyperbolic space to acquire structure- and geometry-aware anchor representations.
Finally, HyperSED builds the partitioning tree of the anchor message graph by incorporating differentiable structural information as the reflection of the detected events.
Extensive experiments on public datasets demonstrate HyperSED's competitive performance, along with a substantial improvement in efficiency compared to the current state-of-the-art unsupervised paradigm.
Statistically, HyperSED boosts incremental SED by 
an average of 2\%, 2\%, and 25\% in NMI, AMI, and ARI, respectively; 
enhancing efficiency by up to 37.41 times and at least 12.10 times, illustrating the advancement of the proposed framework.
Our code is publicly available at \url{https://github.com/XiaoyanWork/HyperSED}.

\end{abstract}

%


\section{Introduction}
    
    Social Event Detection (SED), a task aimed at identify noteworthy occurrences on social media \cite{cordeiro2016online}, remains challenging due to its \textit{large scale}, \textit{complex interrelations}, and \textit{high dynamism} \cite{fedoryszak2019real}. 
    The SED task is pivotal for various downstream applications, such as crisis management \cite{pohl2012automatic}, public opinion monitoring \cite{karamouzas2022public}, etc.
    Current attempts frame SED as a problem of learning representations of social messages and clustering them into events \cite{cao2021knowledge,ren2022known}.

    Recent advancements in modeling social messages into graphs have yielded promising results across supervised \cite{cao2021knowledge}, self-supervised \cite{ren2022known}, and unsupervised \cite{cao2024hierarchical} paradigms. 
    These methods typically construct the social message graph based on message attributes \cite{peng2021streaming} and employ graph neural networks (GNNs), optimized through contrastive learning \cite{cao2021knowledge}, reinforcement learning \cite{peng2022reinforced}, and pseudo-label generation \cite{ren2022known} to learn clustering-friendly message representations. 
    Notably, \citet{cao2024hierarchical} introduced an unsupervised algorithm based on structural entropy \cite{li2016structural} minimization, achieving state-of-the-art performance without any supervision signal.

    However, existing works have neglected some or all of the potential challenges posed by the characteristics of social message data.
    \textbf{Firstly, the large scale of social messages leads to resource-intensive learning.}
        Social messages are often short texts that users post in response to specific events. 
        Among messages describing the same event, certain messages may exhibit high content similarity, even nearly identical semantics.
        When handling such data, current methods \cite{ren2022known,cao2021knowledge} consider all messages for learning, leading to resource wastage.
    \textbf{Secondly, the complex interrelations between} messages require a stereoscopic representation.
        To uncover the relationships among messages, current methods \cite{cao2024hierarchical,ren2023uncertainty} learn message representations within the Euclidean space.
        However, in real-world scenarios, messages often exhibit nested and hierarchical patterns \cite{wang2024metahkg}. 
        When confronted with such data patterns, the Euclidean space lacks the expressiveness required to uncover profound interrelations among them.
    \textbf{Thirdly, the dynamic nature} of novel events continuously emerging necessitates SED systems to react efficiently.
        New events occur continuously, regularly introducing new topics, trends, and relationships, making it costly and laborious to obtain message labels or total event count.
        Efficiently detecting new events from a large-scale message pool without supervision is crucial for SED systems.
        
    To address the aforementioned challenges, this work introduces a novel unsupervised SED framework, HyperSED (\textbf{Hyper}bolic \textbf{SED}).
    Given the vast amount of social messages, we establish the notion of the Semantic-based Anchor Message Graph (SAMG), where semantically related messages are seen as a singular node within the anchor graph.
    To uncover the complex interrelations among messages, we represent the anchor and model the graph structure within the hyperbolic space to derive structure- and geometry-aware anchor representations.
    Finally, we harness the power of differentiable structural information to construct the partitioning tree of the SAMG bottom-up, which formulates the detected events efficiently and effectively.
    Extensive experiments on two publicly available datasets demonstrate HyperSED's efficiency and effectiveness for SED.
    Further evaluation and analyses highlight HyperSED's advancement and applicability as a solid framework.
    In summary, our contributions are as follows:
        
        $\bullet$ We introduce HyperSED, an effective and efficient unsupervised framework for social event detection, which exploits the benefits of graph learning in the hyperbolic space and structural information to achieve event detection.
        
        $\bullet$ We propose a semantic-based anchor message graph, simplifying the vast amount of social messages from a semantic viewpoint and uncovering the complex interrelations among messages within the hyperbolic space.
        
        $\bullet$ Extensive experiments on public datasets demonstrate HyperSED's competitive performance in unsupervised online and offline SED scenarios, marked by significant enhancements in efficiency.

\section{Related Work}

    This section includes the overview of literatures related to Social Event Detection (SED) and Structural Entropy (SE).

    \subsubsection{Social Event Detection}
    
        Social Event Detection (SED) is a long-standing and challenging task \cite{atefeh2015survey}. 
        The vast amount of social messages, covering a wide range of topics \cite{ren2023uncertainty}, poses significant difficulties in learning message representations and categorizing events.
        Earlier SED methods fall in incremental clustering \cite{zhao2007temporal,feng2015streamcube}, topic modeling \cite{xing2016hashtag,wang2016using}, and community detection \cite{fedoryszak2019real,liu2020event,liu2020story}.
        Recent SED methods utilize GNNs' capabilities to model social messages and their relationships \cite{cao2021knowledge,ren2022known,peng2022reinforced,peng2021streaming} by extracting key attributes.
        They employ contrastive learning to pull related messages closer and push unrelated messages apart, optimizing supervised \cite{cao2021knowledge} or self-supervised \cite{ren2022known} SED. 
        However, these methods require supervision signals like a predefined number of events.
        HISEvent, proposed by \citet{cao2024hierarchical}, is an unsupervised SED algorithm that hierarchically minimizes two-dimensional structural entropy, requiring no supervision signals.
        Our work retains the merits of prior approaches and addresses their limitations, achieving effective and efficient unsupervised event detection.

    \subsubsection{Structural Entropy}

        Information entropy \cite{shannon1948mathematical} measures the amount of information in unstructured data but cannot capture information in graph structures. 
        \citet{li2016structural} proposed structural entropy to consider the structural information of graphs. 
        This theory has been applied in various fields, such as graph pooling \cite{wu2022structural}, adversarial attacks \cite{liu2019rem}, contrastive learning \cite{wu2023sega}, and graph structural learning \cite{zou2023se}. 
        In the context of SED, \citet{cao2024hierarchical} proposed minimizing two-dimensional structural entropy to achieve unsupervised event detection. 
        However, these methods utilize the discrete algorithms of this theory and cannot be combined with trainable networks. 
        This limitation was overcome when \citet{sun2024lsenet} reformulated the structural entropy into differentiable structural information for deep graph clustering.
        This has allowed optimizing the partitioning tree proposed by \citet{li2016structural}.

\section{Preliminaries}

    This section outlines the essential concepts and definitions, including Differentiable Structural Information and Hyperbolic Space, and then presents the problem formulation.

    \subsection{Differentiable Structural Information}

        Differentiable Structural Information (DSI, \cite{sun2024lsenet}) is an equivalent reformulation of structural information \cite{li2016structural}.
        This reformulation employs a layer-wise assignment approach, enabling the partitioning tree to be differentiable and thus optimizable. 

        \begin{definition}[Partitioning Tree \cite{li2016structural,cao2024hierarchical}] 
            Given a weighted graph $G=(\mathcal{V}, \mathcal{E})$ with weight function $\omega$, its partitioning tree $\mathcal{T}$ satisfies the following:

            1. For the root node $\lambda$, $T_{\lambda} = \mathcal{V}$.
            For every node $\alpha \in \mathcal{T}$, a subset $T_{\alpha} \subseteq \mathcal{V}$ is associated with it.
            Each leaf node $\gamma$ is associated with a single node in $G$, $T_{\gamma}=\{v\}, v \in \mathcal{V}$.
            
            2. The height of tree $\mathcal{T}$ is $h(\mathcal{T})=H$, $h(\lambda)=H$, $h(\gamma)=0$, and $h(\alpha^-)=h(\alpha)+1$, where $\alpha^-$ denote the parent of $\alpha$.

            3. For each node $\alpha \in \mathcal{T}$, denote its children as $\beta_1, \beta_2, \ldots$; thus, $(T_{\beta_1}, T_{\beta_2}, \ldots)$ is a partition of $T_{\alpha}$. 
        \label{def:par_tree}
        \end{definition}

        \begin{definition}[$H$-Dimensional Structural Information \cite{li2016structural,sun2024lsenet}]
            For a partitioning tree $\mathcal{T}$ with height $H$, the formula of structural information of $G$ at the $h$-th layer of $\mathcal{T}$ is defined as:
            \begin{equation}
                \begin{aligned}
                    \mathcal{H}^\mathcal{T}(G;h)=&-\frac{1}{\operatorname{vol}(G)}\sum\limits_{k=1}^{N_h}(\operatorname{vol}^h(T_k) - \sum\limits_{(i,j)\in \mathcal{E}}S^h_{ik}S^h_{jk}\omega_{ij}) \\
                    & \cdot \log_2\frac{\operatorname{vol}^h(T_k)}{\operatorname{vol}^{h-1}({T_{k^-}})},
                \end{aligned}
            \end{equation}
            where $\operatorname{vol}(\cdot)$ calculate the volume of corresponding node set and $\omega_{ij}$ is the weight between node $i$ and $j$.
            For the $k$-th node in height $h$, $S^h_{ik}=\mathbb{I}(i\in T_{k})$.
            Finally, the $H$-dimensional structural information of $G$ is:
            \begin{equation}
                \mathcal{H}^\mathcal{T}(G) = \sum\nolimits_{h=1}^H H^\mathcal{T}(G;h).
            \label{eq:se}
            \end{equation}
        \end{definition}

    \begin{figure*}[htbp]
        \centering
        \includegraphics[width=1.0\linewidth]{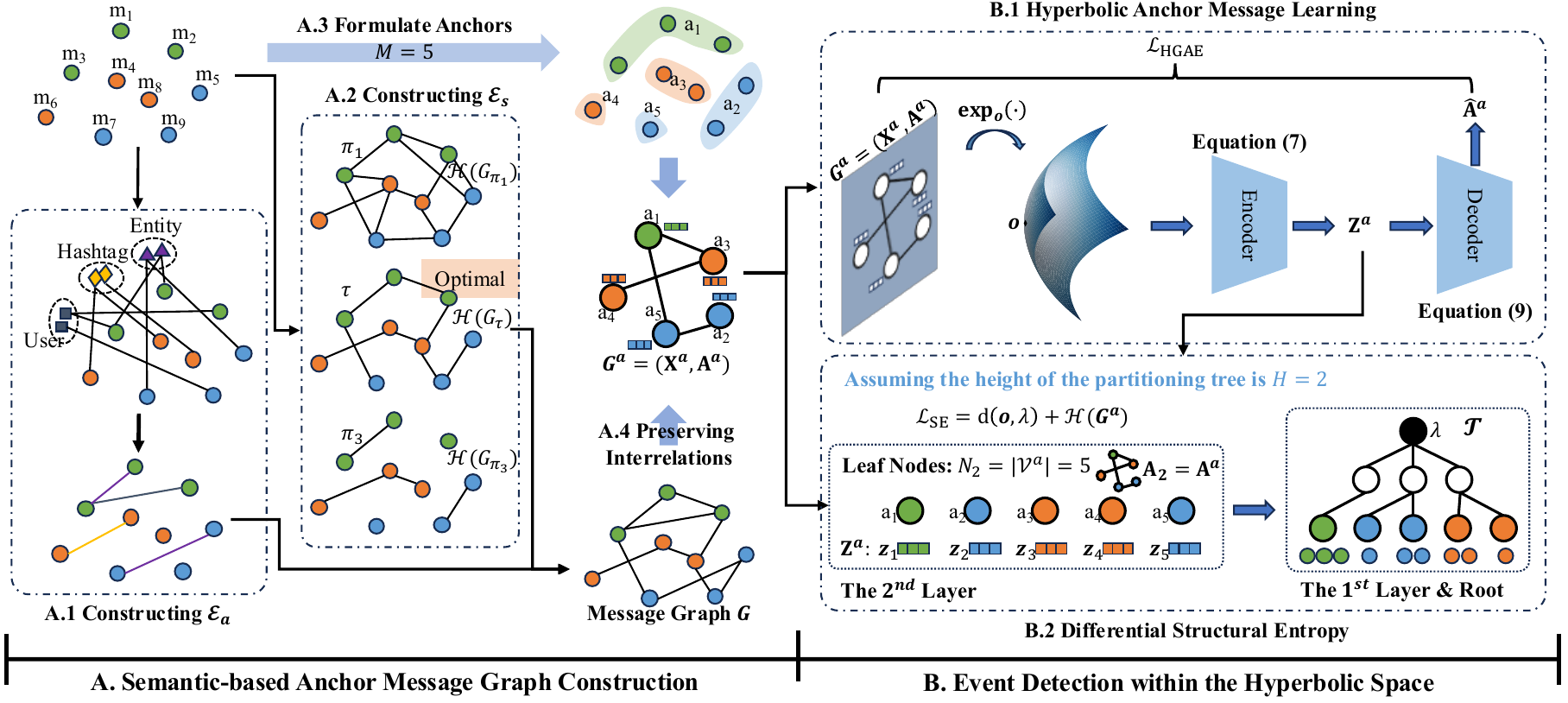}
        \caption{The overall framework of HyperSED.}
        \label{fig:overall_framework}
    \end{figure*}

    \subsection{Hyperbolic Space}
    
        The hyperbolic space, unlike the `flat' Euclidean space, is a curved space with negative curvature $\kappa$, which quantifies the extent to which the manifold deviates from flatness.
        We apply the Poincar\'{e} ball model \cite{ungar2001hyperbolic} of the hyperbolic space, which excels in social network analysis due to its efficient and compact representation of hierarchical and nested structures \cite{wang2024metahkg,papadopoulos2014network}.
        The $d$-dimensional Poincar\'{e} ball model with constant negative curvature $\kappa< 0$ is defined as \cite{ungar2022gyrovector} $\mathbb{B}^d_\kappa = \{\boldsymbol{x} \in \mathbb{R}^d:  \lVert \boldsymbol{x} \rVert^2=-\frac{1}{\kappa}\}$, where $\lVert \cdot \rVert$ denotes Euclidean norm.
        The origin of the Poincar\'{e} ball is $\mathbf{o}=(0,\ldots, 0)\in \mathbb{R}^d$.
        The distance function \cite{ungar2022gyrovector} is given by $d_{\mathbb{B}_\kappa}(\boldsymbol{x}, \boldsymbol{y})=\frac{2}{\sqrt{-\kappa}}\tanh^{-1}(\lVert u \rVert)$, where $u=(-\boldsymbol{x}) \oplus_\kappa \boldsymbol{y}$ and $\oplus_\kappa$ is the M{\"o}bius addition \cite{ungar2001hyperbolic}:
        \begin{equation}
            \boldsymbol{x} \oplus_\kappa \boldsymbol{y} = \frac{(1-2\kappa\boldsymbol{x}^\top\boldsymbol{y}-\kappa\lVert \boldsymbol{y} \rVert^2)\boldsymbol{x}+(1+\kappa\lVert \boldsymbol{x} \rVert^2)\boldsymbol{y}} {1-2\kappa\boldsymbol{x}^\top\boldsymbol{y}+\kappa^2\lVert \boldsymbol{x} \rVert^2\lVert \boldsymbol{y} \rVert^2}.
        \end{equation}
        The exponential and logarithmic maps of Poincar\'{e} ball model are defined as \cite{petersen2006riemannian}:
        \begin{equation}
            \exp_{\boldsymbol{x}}^\kappa(\boldsymbol{v})=\boldsymbol{x} \oplus_\kappa(\frac{\boldsymbol{v}}{\sqrt{-\kappa}\lVert \boldsymbol{v} \rVert}\tanh(\frac{\sqrt{-\kappa}\lambda^\kappa_{\boldsymbol{x}}\lVert \boldsymbol{v} \rVert}{2})),
        \label{eq:exp_map}
        \end{equation}
        \begin{equation}
            \log_{\boldsymbol{x}}^\kappa(\boldsymbol{y})=\frac{2}{\sqrt{-\kappa}\lambda^\kappa_{\boldsymbol{x}}}
            \tanh^{-1}(\sqrt{-\kappa}\lVert u \rVert)\frac{u}{\lVert u \rVert}.
        \end{equation}
        For more details about the hyperbolic space, please refer to the Appendix.

    \subsection{Problem Formulation}

        We technically approach the task of Social Event Detection (SED) as a graph learning and clustering problem and outline the formulation as follows:
        
        \begin{description}
        
            \item[\textbf{Input.}]
            A sequence of social messages $m_1, m_2, ..., \text{and }m_N$, where $N$ represents the total number of messages.
            
            \item[\textbf{Output.}]
            $K$ clusters, each corresponding to a detected event.
            
            \item[\textbf{Goal.}]
            The objective is to construct social message graphs and uncover the correlation between messages to derive an optimal partitioning tree $\mathcal{T}$ without a predetermined number of event clusters.
            The obtained partitioning tree reflects the detected events.
            
        \end{description}

\section{Methodology}

    The proposed framework, HyperSED, as shown in Figure \ref{fig:overall_framework}, comprises two stages: anchor message graph construction and event detection.

    \subsection{Anchor Message Graph Construction}

        We begin by constructing social message graphs from individual messages. 
        Subsequently, an anchor message graph is formulated based on the semantic content of the messages while retaining message interrelations.

        \subsubsection{Message Graph Construction.}
        
            In the message graph $G=(\mathcal{V}, \mathcal{E}_a \cup \mathcal{E}_s)$, we treat all messages as nodes $\mathcal{V}=\{m_1, ..., m_N\}$, and consider two types of edges: attribute edges $\mathcal{E}_a$ and semantic edges $\mathcal{E}_s$.
            Attribute edges convey the co-occurrence of attributes among social messages, such as user $u$, hashtag $h$, and entity $e$ \cite{ren2022known}.
            Semantic edges serve to compensate for potential correlations that may be overlooked due to the absence of co-occurring attributes.
            For each message $m_i$, we extract its attributes $a_i=\{u_i, h_i, e_i\}$. 
            An attribute edge is added in $G$ between $m_i$ and $m_j$ if they share a common attribute, i.e., $\mathcal{E}_a = \{(m_i, m_j) \mid {a_i} \cap {a_j} \neq \emptyset\}$, as shown in Figure \ref{fig:overall_framework}A.1.
            For semantic edges, \citet{cao2024hierarchical} identify $k$ stable neighbors for each node via minimizing the one-dimensional structural entropy (1DSE).
            However, this practice incurs significant computational costs and may not always yield results.
            Therefore, we design an efficient algorithm that determines an optimal similarity threshold $\tau$, preserving edges with weights surpassing it, as shown in Figure \ref{fig:overall_framework}A.2.
            First, all message nodes are interconnected by edge weights corresponding to their similarity.
            The similarity is calculated with cosine similarity between their initial message embeddings (in this work obtained with SBERT \cite{reimers2019sentence}).
            Then, the algorithm computes the 1DSE of graphs across varying thresholds through:
            \begin{equation}
                \mathcal{H}^{(1)}(G)=-\sum_{i=1}^{|V|}\frac{d_i}{\operatorname{vol}(G)}\log\frac{d_i}{\operatorname{vol}(G)}.
            \end{equation}
            The optimal threshold is determined by searches for a relatively stable graph structure and preserving more edges as:
            \begin{equation}
                \tau = \underset{\pi \in \Pi}{\arg\min}\left |\mathcal{H}^{(1)}(G_\pi) -\frac{1}{\lvert \Pi \rvert}\sum_{\pi \in \Pi}\mathcal{H}^{(1)}(G_\pi)\right|,
            \end{equation}
            where $\Pi$ is the search space of thresholds and $\mathcal{H}^{(1)}(G_\pi)$ is the 1DSE of $G$ under the threshold $\pi$.
            The semantic edges are defined as $\mathcal{E}_s=\{(m_i, m_j)|\operatorname{sim}(m_i, m_j)\geq \tau\}$.
            So far, all edges have been established, resulting in the creation of the message graph $G=(\mathcal{V}, \mathcal{E}_a \cup \mathcal{E}_s)=(\mathbf{X}, \mathbf{A})$, where $\mathbf{X}$ is the embedding matrix and $\mathbf{A}$ denotes the adjacency matrix with $A_{ij}=\max(\operatorname{sim}(\boldsymbol{x}_i, \boldsymbol{x}_j), 0), \text{s.t.} (m_i, m_j) \in \mathcal{E}_a \cup \mathcal{E}_s$.

        \subsubsection{Anchor Message Graph Construction.}
    
            With a substantial size of social messages covering diverse topics, many exhibit semantic similarities or near-identical content. 
            Current methods for SED generally consider all messages within the message graph \cite{cao2024hierarchical,ren2023uncertainty}, which makes the graph structure complex and the learning process resource-intensive.
            This work addresses this by constructing a Semantic-based Anchor Message Graph (\textbf{SAMG}), which regards semantically similar messages as a single anchor node while preserving their relationships through edges.
            The essence of constructing the anchor graph is to reduce the size of the message graph while preserving the integrity of the information conveyed by the original messages.
            To achieve this goal, we derive $M$ anchor nodes, denoted as $\{a_1, \ldots, a_M\}$, based on the semantic similarities between message embeddings $\mathbf{X}$. 
            Each anchor comprises a set of semantically related messages, illustrated in Figure \ref{fig:overall_framework}A.3.
            The embedding $\boldsymbol{x}^a_u$ of the anchor node $a_u$ is computed as the arithmetic mean of the message representations within the anchor, given by $\boldsymbol{x}^a_u=\frac{1}{|a_u|}\sum \{\boldsymbol{x}_{i}|m_i \in a_u\}$.
            The adjacency matrix $\mathbf{A}^a$ of the anchor graph is defined as $\mathbf{A}^a = (\mathbf{C})^\top\mathbf{A}\mathbf{C}$, where $\mathbf{C} \in \mathbb{R}^{N \times M}$ and $C_{iu}=\mathbb{I}(m_i \in a_u)$.
            It signifies that the edge weight in the anchor graph preserves the connections between messages not assigned to the same anchor while accounting for the interrelations of messages from an anchor perspective. 
            Thus, the SAMG $G_a=(\mathbf{X}^a, \mathbf{A}^a)$ is obtained, with each node serving as a representative of a group of semantically related messages and edges as the correlations between the messages of different anchors.
            
            \begin{remark}
                SBERT effectively captures message semantics.
                By applying clustering techniques to these representations, the resulting anchors are believed to group semantically similar messages effectively.
            \end{remark}

    \subsection{Event Detection within the Hyperbolic Space}

        Given the SAMG, we take a three-step effort to achieve unsupervised social event detection, as shown in Figure \ref{fig:overall_framework}B.
        The primary objective is to capture intricate relationships among anchors and obtain the optimal partitioning tree of the graph.

            \textit{\textbf{Step 1}: Learning the SAMG with \textbf{graph autoencoder in the hyperbolic space (HGAE)}.}
            This step aims to acquire structure- and geometry-aware anchor representations by learning the anchor graph in the hyperbolic space.
            The structure-aware representation captures the relational dependencies and patterns among anchors.
            The geometry-aware representation reflects the spatial and topological correlations among anchors within the hyperbolic space, indicating their stereoscopic distances.
            To achieve this, we first define the hyperbolic graph autoencoder (\textbf{HGAE}).
            Inspired by prior works \cite{kipf2016variational,chami2019hyperbolic,park2021unsupervised,sun2024lsenet}, we first map the Euclidean anchor ($a_i$) embedding $\boldsymbol{x}^a_i\in \mathbb{R}^d$ into the hyperbolic space by $\boldsymbol{h}_i = \exp_{\mathbf{o}}^\kappa(\boldsymbol{x}^a_i)\in \mathbb{B}_\kappa^d$ (Equation \ref{eq:exp_map}), where $d$ is the embedding's dimension.
            The graph encoder in the hyperbolic space, which is basically a hyperbolic convolutional layer \cite{chami2019hyperbolic}, is defined as:
            \begin{equation}
                \boldsymbol{z}_i = \operatorname{PConv}(\boldsymbol{h}_i)=\exp_{\mathbf{o}}^{\kappa}\left(\sum_{j=1}^N\omega_{ij}\left(\mathbf{W}\log_{\mathbf{o}}^\kappa(\boldsymbol{h}_i)+b\right)\right),
            \end{equation}
            where $\boldsymbol{z}_i$ is the latent representation of the anchor $a_i$ and $\omega_{ij}$ is derived from the attention mechanism \cite{vaswani2017attention}:
            \begin{equation}
                    \omega_{ij}=\frac{\exp\left(-\frac{1}{\sqrt{|M|}} d_{\mathbb{B}}^2(\boldsymbol{h}_i, \boldsymbol{h}_j)\right)}{\sum_{j=1}^M \exp\left(-\frac{1}{\sqrt{|M|}}d_{\mathbb{B}}^2(\boldsymbol{h}_i, \boldsymbol{h}_{j})\right)}.
            \end{equation}
            The decoder, which reconstructs the adjacency matrix, defines the reconstruction in the hyperbolic space using the Fermi-Dirac distribution \cite{krioukov2010hyperbolic,park2021unsupervised}:
            \begin{equation}
                \hat{A}_{ij}=\sigma\left(\exp(\frac{d^2_{\mathbb{B}}(\boldsymbol{z}_i,\boldsymbol{z}_j)-q}{t})+1\right)^{-1},
            \end{equation}
            where $\hat{\mathbf{A}}$ is the reconstructed anchor adjacency matrix, $\sigma$ is the activation function, and $q$ and $t$ are hyperparameters.
            Finally, the reconstruction loss of the affinity matrix is \cite{kipf2016variational}:
            \begin{equation}
                \mathcal{L}_\text{HGAE}= \mathbb{E}_{q(\mathbf{Z} \mid \mathbf{H}, \mathbf{A}^a)}[\log p(\hat{\mathbf{A}}^a \mid \mathbf{Z})].
            \end{equation}

        \textit{\textbf{Step 2: Bottom-up Learning of Tree Node Embeddings.}}
            This step seeks to build a partitioning tree of the SAMG.
            Following Definition \ref{def:par_tree}, where each graph node corresponds to a leaf node in the partitioning tree, the partitioning tree can be built by iteratively assigning parent nodes layer-by-layer, starting from the leaf nodes \cite{sun2024lsenet}.
            DSI designs an assignment scheme that assigns the $i$-th node on the $h$-th layer to the $j$-th node on the $(h-1)$-th layer by:
            \begin{equation}
                \mathbf{C}^h=\operatorname{Softmax}\left(\mathbf{A}^h\sigma\left(\operatorname{PConv}(\mathbf{Z}^h_a)\right)\right)\in \mathbb{R}^{{N_h}\times N_{h-1}},
            \label{eq:ass}
            \end{equation}
            where ${N_h}$ and $N_{h-1}$ are the number of nodes on the $h$-th and the $(h-1)$-th layer, respectively.
            $\operatorname{Softmax}$ is for row normalization.
            Upon establishing correspondence, the embedding and adjacency matrix of the $(h-1)$-th layer is required, as the construction of the $(h-2)$-th layer relies on this information (as implied by Equation \ref{eq:ass}).
            Similar to building the anchor graph, the embedding of the nodes on the $(h-1)$-th layer is defined with the arithmetic mean of all the embeddings of nodes assigned to it. 
            Only in this case, in the hyperbolic space, the arithmetic mean can be seen as finding the point in the hyperbolic space that is the closest to all nodes.
            This process can be formalized as the Fr\'{e}chet mean \cite{lou2020differentiating}:
            \begin{equation}
                \boldsymbol z^{h-1}_j=\underset{\boldsymbol z^{h-1}_j}{\arg\min} \sum\nolimits_{i=1}^{N^h} c_{ij}d^2_{\mathbb{B}}(\boldsymbol z^{h-1}_j, \boldsymbol z^h_i).
                \label{eq:rep_mean}
            \end{equation}
            The adjacency matrix of nodes on the $(h-1)$-th layer can be derived from it on the $h$-th layer, that is $\mathbf{A}^{h-1} = (\mathbf{C}^h)^\top\hat{\mathbf{A}^h} \mathbf{C}^h$.
            We recursively utilize Equation \ref{eq:ass} and Equation \ref{eq:rep_mean} to build the partitioning tree from leaf nodes to the root node in hyperbolic space.

        \textit{\textbf{Step 3: Optimization of the partitioning tree.}}
            In essence, the optimization objective encompasses two primary aspects: first, find the proper position to put the root node of the partitioning tree, and second, minimize the structural entropy of the SAMG.
            For the first aspect, a straightforward approach is to position the root node at the origin $\mathbf{o}$ of the Poincaré ball. 
            This choice is significant as the origin serves as a reference point for mapping from Euclidean to hyperbolic space, where all other points exhibit symmetric positions relative to it.
            Regarding the second aspect, which involves minimizing the H-dimensional structural entropy corresponding to Equation \ref{eq:se}.
            The loss function is given by:
            \begin{equation}
                \mathcal{L}_\text{SE} = d_{\mathbb{B}}(\boldsymbol{o}, \boldsymbol{z}^0) + \sum\nolimits_{h=1}^H H^\mathcal{T}(G;h),
            \end{equation}
            where $\boldsymbol{z}^0$ is the embedding of the root node.
            The total loss of the training is $\mathcal{L}=\mathcal{L}_\text{HGAE}+\mathcal{L}_\text{SE}$.

        \begin{remark}
            The overall methodology can be seen as starting from individual messages, layer-by-layer, to the top of the partitioning tree. 
            Only the immediate parent layer of messages is chosen based on their semantic relatedness rather than the layer-wise assignment and the value of structural entropy.
            This strategy maintains the benefits of SE while notably enhancing efficiency through learning with anchors.
        \end{remark}

    \subsection{Time Complexity of HyperSED}

        First, the time complexity for the message graph construction including construct attribute edges: $O(|\mathcal{E}_a|)$, calculating the similarity between messages: $O(N^2)$, and constructing the semantic edges: $O(|\Pi|\cdot|\mathcal{E}_s|)$.
        Second, constructing the SAMG has the time complexity of $O(N)$.
        Third, the DSI is computed level-wisely, so the time complexity at level $h$ is $O(N_h|\mathcal{E}|)$.
        Fourth, the time complexity for constructing the partitioning tree is $O(HM)$.
        Therefore, the overall time complexity comes to $O(|\mathcal{E}_a|+N^2 + N + M|\mathcal{E}|)$, as $N_h \ll M$, $H$ denote the height of $\mathcal{T}$, and $|\Pi|$ is the search size of threshold, which are small constant numbers.

\section{Experiments}
        
    We conduct experiments and analyses to evaluate the \textbf{effectiveness} and \textbf{efficiency} of the proposed framework. 
    Additionally, we present \textit{ablation studies} and \textit{parameter analyses} to demonstrate the advancement of the proposed framework.
        
        \begin{table}[htbp]
        \centering
        \caption{Offline results. The best result is marked in \textbf{bold}, the runner-up is marked in \underline{\textit{italic}}, `*' marks results acquired with the ground truth event numbers, the proposed method is marked in \colorbox[HTML]{FFCE93}{orange} background.}
        \resizebox{1.0\linewidth}{!}{
        \begin{tabular}{l|ccc|ccc}
        \hline
        Datasets  & \multicolumn{3}{c|}{English Twitter}                                                & \multicolumn{3}{c}{French Twitter}                                                \\
        \hline
        Metrics   & NMI                         & AMI                       & ARI                       & NMI                       & AMI                       & ARI                       \\
        \hline
        BERT*     & 0.42                        & 0.11                      & 0.01                      & 0.26                      & 0.10                      & 0.01                      \\
        SBERT*    & \underline{\textit{0.83}}   & \underline{\textit{0.73}} & 0.17                      & \underline{\textit{0.66}} & \underline{\textit{0.59}} & 0.10                      \\
        EventX    & 0.72                        & 0.19                      & 0.05                      & 0.56                      & 0.16                      & 0.03                      \\
        \hline
        KPGNN*    & 0.79                        & 0.52                      & \underline{\textit{0.22}} & 0.56                      & 0.44                      & 0.15                      \\
        QSGNN*    & 0.72                        & 0.53                      & \underline{\textit{0.22}} & 0.58                      & 0.44                      & 0.16                      \\
        HISEvent  & 0.47                        & 0.42                      & 0.17                      & 0.48                      & 0.41                      & \underline{\textit{0.30}} \\
        \hline
        \rowcolor[HTML]{FFCE93} 
        HyperSED  & \textbf{0.84}               & \textbf{0.77}             & \textbf{0.44}             & \textbf{0.68}             & \textbf{0.62}             & \textbf{0.62}             \\
        promotion & $\uparrow$ 0.01             & $\uparrow$ 0.04           & $\uparrow$ 0.22           & $\uparrow$  0.02          & $\uparrow$  0.03          &  $\uparrow$  0.32         \\
        \hline
        \end{tabular}
        }
        \label{tab:offline}
        \end{table}

    \subsection{Experimental Setups}
    
        \begin{table*}[!t]
        \centering
        \caption{Online results of Twitter English. The best result is marked in \textbf{bold}, the runner-up is marked in \underline{\textit{italic}}, `*' marks results acquired with the ground truth event numbers, the proposed method is marked in \colorbox[HTML]{FFCE93}{orange} background.}
        \resizebox{1\linewidth}{!}{
        \begin{tabular}{l|ccc|ccc|ccc|ccc|ccc|ccc|ccc}
        \hline
        Blocks    & \multicolumn{3}{c|}{$\text{M}_{1}$}                                                 & \multicolumn{3}{c|}{$\text{M}_{2}$}                                               & \multicolumn{3}{c|}{$\text{M}_{3}$}                                               & \multicolumn{3}{c|}{$\text{M}_{4}$}                                               & \multicolumn{3}{c|}{$\text{M}_{5}$}                                               & \multicolumn{3}{c|}{$\text{M}_{6}$}                                               & \multicolumn{3}{c}{$\text{M}_{7}$}                                                \\
        \hline
        Metrics   & NMI                         & AMI                       & ARI                       & NMI                       & AMI                       & ARI                       & NMI                       & AMI                       & ARI                       & NMI                       & AMI                       & ARI                       & NMI                       & AMI                       & ARI                       & NMI                       & AMI                       & ARI                       & NMI                       & AMI                       & ARI                       \\
        \hline
        BERT*     & 0.14                        & 0.11                      & 0.01                      & 0.38                      & 0.32                      & 0.43                      & 0.22                      & 0.15                      & 0.05                      & 0.26                      & 0.18                      & 0.03                      & 0.28                      & 0.23                      & 0.08                      & 0.32                      & 0.20                      & 0.08                      & 0.20                      & 0.14                      & 0.01                      \\
        SBERT*    & 0.40                        & 0.38                      & 0.03                      & \underline{\textit{0.85}} & \underline{\textit{0.84}} & 0.70                      & \underline{\textit{0.89}} & \underline{\textit{0.88}} & 0.67                      & \textbf{0.80}             & \textbf{0.78}             & 0.36                      & \textbf{0.87}             & \textbf{0.87}             & \textbf{0.79}             & \underline{\textit{0.86}} & \underline{\textit{0.83}} & 0.56                      & 0.63                      & 0.60                      & 0.09                      \\
        EventX    & 0.36                        & 0.06                      & 0.01                      & 0.68                      & 0.29                      & 0.45                      & 0.63                      & 0.18                      & 0.09                      & 0.63                      & 0.19                      & 0.07                      & 0.59                      & 0.14                      & 0.04                      & 0.70                      & 0.27                      & 0.14                      & 0.51                      & 0.13                      & 0.02                      \\
        \hline
        KPGNN*    & 0.39                        & 0.37                      & 0.07                      & 0.79                      & 0.78                      & 0.76                      & 0.76                      & 0.74                      & 0.58                      & 0.67                      & 0.64                      & 0.29                      & 0.73                      & 0.71                      & 0.47                      & 0.82                      & 0.79                      & 0.72                      & 0.55                      & 0.51                      & 0.12                      \\
        QSGNN*    & \underline{\textit{0.43}}   & \underline{\textit{0.41}} & 0.07                      & 0.81                      & 0.80                      & \underline{\textit{0.77}} & 0.78                      & 0.76                      & 0.59                      & 0.71                      & 0.68                      & 0.29                      & 0.75                      & 0.73                      & 0.48                      & 0.83                      & 0.80                      & 0.73                      & 0.57                      & 0.54                      & 0.12                      \\
        HISEvent  & 0.38                        & 0.37                      & \underline{\textit{0.09}} & \textbf{0.90}             & \textbf{0.89}             & \textbf{0.88}             & \textbf{0.90}             & \textbf{0.89}             & \underline{\textit{0.79}} & \underline{\textit{0.77}} & \underline{\textit{0.76}} & \underline{\textit{0.52}} & 0.83                      & \underline{\textit{0.82}} & 0.63                      & \textbf{0.89}             & \textbf{0.88}             & \textbf{0.84}             & \underline{\textit{0.64}} & \underline{\textit{0.63}} & \underline{\textit{0.36}} \\
        \hline
        \rowcolor[HTML]{FFCE93} 
        HyperSED  & \textbf{0.84}               & \textbf{0.84}             & \textbf{0.96}             & 0.84                      & 0.83                      & \underline{\textit{0.77}} & 0.85                      & 0.84                      & \textbf{0.81}             & \underline{\textit{0.77}} & 0.75                      & \textbf{0.60}             & \underline{\textit{0.84}} & \underline{\textit{0.82}} & \underline{\textit{0.71}} & \underline{\textit{0.86}} & \underline{\textit{0.83}} & \underline{\textit{0.76}} & \textbf{0.85}             & \textbf{0.85}             & \textbf{0.89}             \\
        promotion & $\uparrow$ 0.41             & $\uparrow$ 0.43           & $\uparrow$ 0.87           & $\downarrow$ 0.06         & $\downarrow$ 0.06         & $\downarrow$ 0.11         & $\downarrow$ 0.05         & $\downarrow$ 0.05         & $\uparrow$ 0.02           & $\downarrow$ 0.03         & $\downarrow$ 0.03         & $\uparrow$ 0.08           & $\downarrow$ 0.03         & $\downarrow$ 0.05         & $\downarrow$ 0.08         & $\downarrow$ 0.03         & $\downarrow$ 0.05         & $\downarrow$ 0.08         & $\uparrow$ 0.21           & $\uparrow$ 0.22           & $\uparrow$ 0.53           \\
        \hline\hline
        Blocks    & \multicolumn{3}{c|}{$\text{M}_{8}$}                                                 & \multicolumn{3}{c|}{$\text{M}_{9}$}                                               & \multicolumn{3}{c|}{$\text{M}_{10}$}                                              & \multicolumn{3}{c|}{$\text{M}_{11}$}                                              & \multicolumn{3}{c|}{$\text{M}_{12}$}                                              & \multicolumn{3}{c|}{$\text{M}_{13}$}                                              & \multicolumn{3}{c}{$\text{M}_{14}$}                                               \\
        \hline
        Metrics   & NMI                         & AMI                       & ARI                       & NMI                       & AMI                       & ARI                       & NMI                       & AMI                       & ARI                       & NMI                       & AMI                       & ARI                       & NMI                       & AMI                       & ARI                       & NMI                       & AMI                       & ARI                       & NMI                       & AMI                       & ARI                       \\
        \hline
        BERT*     & 0.29                        & 0.15                      & 0.04                      & 0.30                      & 0.20                      & 0.09                      & 0.30                      & 0.20                      & 0.07                      & 0.29                      & 0.20                      & 0.06                      & 0.20                      & 0.15                      & 0.04                      & 0.25                      & 0.18                      & 0.04                      & 0.22                      & 0.16                      & 0.05                      \\
        SBERT*    & \textbf{0.87}               & \textbf{0.85}             & 0.62                      & \underline{\textit{0.86}} & \underline{\textit{0.84}} & 0.47                      & \underline{\textit{0.88}} & \underline{\textit{0.86}} & 0.61                      & 0.81                      & \underline{\textit{0.79}} & 0.37                      & \underline{\textit{0.84}} & \underline{\textit{0.83}} & 0.60                      & 0.72                      & 0.69                      & 0.21                      & 0.77                      & 0.75                      & 0.36                      \\
        EventX    & 0.71                        & 0.21                      & 0.09                      & 0.67                      & 0.19                      & 0.07                      & 0.68                      & 0.24                      & 0.13                      & 0.65                      & 0.24                      & 0.16                      & 0.61                      & 0.16                      & 0.07                      & 0.58                      & 0.16                      & 0.04                      & 0.57                      & 0.14                      & 0.10                      \\
        \hline
        KPGNN*    & 0.80                        & 0.76                      & 0.60                      & 0.74                      & 0.71                      & 0.46                      & 0.80                      & 0.78                      & 0.70                      & 0.74                      & 0.71                      & 0.49                      & 0.68                      & 0.66                      & 0.48                      & 0.69                      & 0.67                      & 0.29                      & 0.69                      & 0.65                      & 0.42                      \\
        QSGNN*    & 0.79                        & 0.75                      & 0.59                      & 0.77                      & 0.75                      & 0.47                      & 0.82                      & 0.80                      & 0.71                      & 0.75                      & 0.72                      & 0.49                      & 0.70                      & 0.68                      & 0.49                      & 0.68                      & 0.66                      & 0.29                      & 0.68                      & 0.66                      & 0.41                      \\
        HISEvent  & \underline{\textit{0.82}}   & \underline{\textit{0.81}} & \underline{\textit{0.68}} & \textbf{0.89}             & \textbf{0.88}             & \underline{\textit{0.65}} & \textbf{0.91}             & \textbf{0.90}             & \textbf{0.87}             & \underline{\textit{0.85}} & \textbf{0.84}             & \underline{\textit{0.66}} & \textbf{0.87}             & \textbf{0.87}             & \textbf{0.82}             & \underline{\textit{0.75}} & \underline{\textit{0.74}} & \underline{\textit{0.39}} & \underline{\textit{0.83}} & \underline{\textit{0.82}} & \underline{\textit{0.71}} \\
        \hline
        \rowcolor[HTML]{FFCE93} 
        HyperSED  & \textbf{0.87}               & \textbf{0.85}             & \textbf{0.69}             & \underline{\textit{0.86}} & \underline{\textit{0.84}} & \textbf{0.81}             & 0.86                      & 0.85                      & \underline{\textit{0.74}} & \textbf{0.86}             & \textbf{0.84}             & \textbf{0.91}             & 0.78                      & 0.76                      & \underline{\textit{0.62}} & \textbf{0.82}             & \textbf{0.81}             & \textbf{0.93}             & \textbf{0.84}             & \textbf{0.83}             & \textbf{0.80}             \\
        promotion & 0.00                        & 0.00                      & $\uparrow$ 0.01           & $\downarrow$ 0.03         & $\downarrow$ 0.04         & $\uparrow$ 0.16           & $\downarrow$ 0.05         & $\downarrow$ 0.05         & $\downarrow$ 0.13         & $\uparrow$ 0.01           & 0.00                      & $\uparrow$ 0.25           & $\downarrow$ 0.09         & $\downarrow$ 0.11         & $\downarrow$ 0.20         & $\uparrow$ 0.07           & $\uparrow$ 0.07           & $\uparrow$ 0.54           & $\uparrow$ 0.01           & $\uparrow$ 0.01           & $\uparrow$ 0.09           \\
        \hline\hline
        Blocks    & \multicolumn{3}{c|}{$\text{M}_{15}$}                                                & \multicolumn{3}{c|}{$\text{M}_{16}$}                                              & \multicolumn{3}{c|}{$\text{M}_{17}$}                                              & \multicolumn{3}{c|}{$\text{M}_{18}$}                                              & \multicolumn{3}{c|}{$\text{M}_{19}$}                                              & \multicolumn{3}{c|}{$\text{M}_{20}$}                                              & \multicolumn{3}{c}{$\text{M}_{21}$}                                               \\
        \hline
        Metrics   & NMI                         & AMI                       & ARI                       & NMI                       & AMI                       & ARI                       & NMI                       & AMI                       & ARI                       & NMI                       & AMI                       & ARI                       & NMI                       & AMI                       & ARI                       & NMI                       & AMI                       & ARI                       & NMI                       & AMI                       & ARI                       \\
        \hline
        BERT*     & 0.24                        & 0.17                      & 0.03                      & 0.28                      & 0.19                      & 0.07                      & 0.22                      & 0.16                      & 0.04                      & 0.20                      & 0.14                      & 0.04                      & 0.28                      & 0.22                      & 0.08                      & 0.32                      & 0.19                      & 0.07                      & 0.20                      & 0.15                      & 0.03                      \\
        SBERT*    & \underline{\textit{0.71}}   & \underline{\textit{0.68}} & 0.17                      & 0.80                      & 0.77                      & 0.50                      & \underline{\textit{0.79}} & \underline{\textit{0.78}} & 0.42                      & \textbf{0.81}             & \textbf{0.80}             & 0.52                      & \textbf{0.86}             & \textbf{0.85}             & 0.56                      & \textbf{0.84}             & \textbf{0.80}             & 0.54                      & 0.72                      & 0.70                      & 0.26                      \\
        EventX    & 0.49                        & 0.07                      & 0.01                      & 0.62                      & 0.19                      & 0.08                      & 0.58                      & 0.18                      & 0.12                      & 0.59                      & 0.16                      & 0.08                      & 0.60                      & 0.16                      & 0.07                      & 0.67                      & 0.18                      & 0.11                      & 0.53                      & 0.10                      & 0.01                      \\
        \hline
        KPGNN*    & 0.58                        & 0.54                      & 0.17                      & 0.79                      & 0.77                      & 0.66                      & 0.70                      & 0.68                      & 0.43                      & 0.68                      & 0.66                      & 0.47                      & 0.73                      & 0.71                      & 0.51                      & 0.72                      & 0.68                      & 0.51                      & 0.60                      & 0.57                      & 0.20                      \\
        QSGNN*    & 0.59                        & 0.55                      & 0.17                      & 0.78                      & 0.76                      & 0.65                      & 0.71                      & 0.69                      & 0.44                      & 0.70                      & 0.68                      & 0.48                      & 0.73                      & 0.70                      & 0.50                      & 0.73                      & \underline{\textit{0.69}} & 0.51                      & 0.61                      & 0.58                      & 0.21                      \\
        HISEvent  & 0.69                        & 0.67                      & \underline{\textit{0.27}} & \underline{\textit{0.87}} & \underline{\textit{0.86}} & \underline{\textit{0.83}} & 0.77                      & 0.76                      & \underline{\textit{0.56}} & 0.74                      & 0.73                      & \underline{\textit{0.64}} & \underline{\textit{0.85}} & \underline{\textit{0.84}} & \underline{\textit{0.60}} & 0.82                      & \textbf{0.80}             & \textbf{0.67}             & \underline{\textit{0.73}} & \underline{\textit{0.73}} & \underline{\textit{0.46}} \\
        \hline
        \rowcolor[HTML]{FFCE93} 
        HyperSED  & \textbf{0.81}               & \textbf{0.80}             & \textbf{0.93}             & \textbf{0.90}             & \textbf{0.89}             & \textbf{0.88}             & \textbf{0.83}             & \textbf{0.82}             & \textbf{0.89}             & \underline{\textit{0.80}} & \underline{\textit{0.79}} & \textbf{0.71}             & \textbf{0.86}             & \textbf{0.85}             & \textbf{0.84}             & \underline{\textit{0.83}} & \textbf{0.80}             & \underline{\textit{0.66}} & \textbf{0.80}             & \textbf{0.78}             & \textbf{0.86}             \\
        promotion & $\uparrow$ 0.10             & $\uparrow$ 0.12           & $\uparrow$ 0.66           & $\uparrow$ 0.03           & $\uparrow$ 0.03           & $\uparrow$ 0.05           & $\uparrow$ 0.04           & $\uparrow$ 0.04           & $\uparrow$ 0.33           & $\downarrow$ 0.01         & $\downarrow$ 0.01         & $\uparrow$ 0.07           & 0.00                      & 0.00                      & $\uparrow$ 0.24           & $\downarrow$ 0.01         & 0.00                      & $\downarrow$ 0.01         & $\uparrow$ 0.07           & $\uparrow$ 0.05           & $\uparrow$ 0.40           \\
        \hline
        \end{tabular}
        }
        \label{tab:online_en}
        \end{table*}

        \begin{table*}[t]
        \centering
        \caption{Online results of Twitter French. The best result is marked in \textbf{bold}, the runner-up is marked in \underline{\textit{italic}}, `*' marks results acquired with the ground truth event numbers, the proposed method is marked in \colorbox[HTML]{FFCE93}{orange} background.}
        \resizebox{1\linewidth}{!}{
        \begin{tabular}{l|ccc|ccc|ccc|ccc|ccc|ccc|ccc|ccc}
        \hline
        Blocks    & \multicolumn{3}{c|}{$\text{M}_{1}$}                                                 & \multicolumn{3}{c|}{$\text{M}_{2}$}                                               & \multicolumn{3}{c|}{$\text{M}_{3}$}                                               & \multicolumn{3}{c|}{$\text{M}_{4}$}                                               & \multicolumn{3}{c|}{$\text{M}_{5}$}                                               & \multicolumn{3}{c|}{$\text{M}_{6}$}                                               & \multicolumn{3}{c|}{$\text{M}_{7}$}                                               & \multicolumn{3}{c}{$\text{M}_{8}$}                                                \\
        \hline
        Metrics   & NMI                         & AMI                       & ARI                       & NMI                       & AMI                       & ARI                       & NMI                       & AMI                       & ARI                       & NMI                       & AMI                       & ARI                       & NMI                       & AMI                       & ARI                       & NMI                       & AMI                       & ARI                       & NMI                       & AMI                       & ARI                       & NMI                       & AMI                       & ARI                       \\
        \hline
        BERT*     & 0.13                        & 0.11                      & 0.04                      & 0.11                      & 0.09                      & 0.04                      & 0.07                      & 0.05                      & 0.01                      & 0.07                      & 0.05                      & 0.01                      & 0.12                      & 0.08                      & 0.02                      & 0.09                      & 0.06                      & 0.01                      & 0.14                      & 0.12                      & 0.05                      & 0.14                      & 0.10                      & 0.04                      \\
        SBERT*    & 0.59                        & 0.58                      & 0.21                      & 0.64                      & 0.63                      & 0.32                      & 0.64                      & 0.64                      & 0.32                      & 0.58                      & 0.57                      & 0.21                      & \textbf{0.74}             & \textbf{0.73}             & 0.41                      & 0.70                      & 0.69                      & 0.36                      & \underline{\textit{0.67}} & \underline{\textit{0.66}} & 0.29                      & 0.76                      & 0.75                      & 0.48                      \\
        EventX    & 0.34                        & 0.11                      & 0.02                      & 0.37                      & 0.12                      & 0.02                      & 0.39                      & 0.01                      & 0.11                      & 0.39                      & 0.14                      & 0.06                      & 0.53                      & 0.24                      & 0.13                      & 0.44                      & 0.15                      & 0.08                      & 0.41                      & 0.12                      & 0.02                      & 0.54                      & 0.21                      & 0.09                      \\
        \hline
        KPGNN*    & 0.54                        & 0.54                      & 0.17                      & 0.56                      & 0.55                      & 0.18                      & 0.55                      & 0.15                      & 0.55                      & 0.55                      & 0.55                      & 0.17                      & 0.58                      & 0.57                      & 0.21                      & 0.59                      & 0.57                      & 0.21                      & 0.63                      & 0.61                      & 0.30                      & 0.58                      & 0.57                      & 0.20                      \\
        QSGNN*    & 0.57                        & 0.56                      & 0.18                      & 0.58                      & 0.57                      & 0.19                      & 0.58                      & 0.17                      & \underline{\textit{0.56}} & 0.58                      & 0.57                      & 0.18                      & 0.61                      & 0.59                      & 0.23                      & 0.60                      & 0.59                      & 0.21                      & 0.64                      & 0.63                      & 0.30                      & 0.57                      & 0.55                      & 0.19                      \\
        HISEvent  & \underline{\textit{0.74}}   & \underline{\textit{0.74}} & \underline{\textit{0.58}} & \underline{\textit{0.73}} & \underline{\textit{0.73}} & \underline{\textit{0.60}} & \textbf{0.72}             & \textbf{0.72}             & 0.52                      & \underline{\textit{0.67}} & \underline{\textit{0.66}} & \underline{\textit{0.48}} & \textbf{0.74}             & \textbf{0.73}             & \underline{\textit{0.56}} & \textbf{0.80}             & \textbf{0.79}             & \underline{\textit{0.66}} & \textbf{0.79}             & \textbf{0.78}             & \underline{\textit{0.59}} & \textbf{0.82}             & \textbf{0.82}             & \textbf{0.75}             \\
        \hline
        \rowcolor[HTML]{FFCE93} 
        HyperSED  & \textbf{0.77}               & \textbf{0.77}             & \textbf{0.89}             & \textbf{0.75}             & \textbf{0.75}             & \textbf{0.77}             & \underline{\textit{0.71}} & \underline{\textit{0.71}} & \textbf{0.81}             & \textbf{0.70}             & \textbf{0.70}             & \textbf{0.59}             & \underline{\textit{0.68}} & \underline{\textit{0.65}} & \textbf{0.57}             & \underline{\textit{0.78}} & \underline{\textit{0.77}} & \textbf{0.86}             & \underline{\textit{0.67}} & \underline{\textit{0.66}} & \textbf{0.75}             & \underline{\textit{0.79}} & \underline{\textit{0.78}} & \underline{\textit{0.64}} \\
        promotion & $\uparrow$ 0.03             & $\uparrow$ 0.03           & $\uparrow$ 0.31           & $\uparrow$ 0.02           & $\uparrow$ 0.02           & $\uparrow$ 0.17           & $\downarrow$ 0.01         & $\downarrow$ 0.01         & $\uparrow$ 0.25           & $\uparrow$ 0.03           & $\uparrow$ 0.04           & $\uparrow$ 0.11           & $\downarrow$ 0.06         & $\downarrow$ 0.08         & $\uparrow$  0.01          & $\downarrow$ 0.02         & $\downarrow$ 0.02         & $\uparrow$ 0.20           & $\downarrow$ 0.12         & $\downarrow$ 0.12         & $\uparrow$ 0.16           & $\downarrow$ 0.03         & $\downarrow$ 0.04         & $\downarrow$ 0.11         \\
        \hline\hline
        Blocks    & \multicolumn{3}{c|}{$\text{M}_{9}$}                                                 & \multicolumn{3}{c|}{$\text{M}_{10}$}                                              & \multicolumn{3}{c|}{$\text{M}_{11}$}                                              & \multicolumn{3}{c|}{$\text{M}_{12}$}                                              & \multicolumn{3}{c|}{$\text{M}_{13}$}                                              & \multicolumn{3}{c|}{$\text{M}_{14}$}                                              & \multicolumn{3}{c|}{$\text{M}_{15}$}                                              & \multicolumn{3}{c}{$\text{M}_{16}$}                                               \\
        \hline
        Metrics   & NMI                         & AMI                       & ARI                       & NMI                       & AMI                       & ARI                       & NMI                       & AMI                       & ARI                       & NMI                       & AMI                       & ARI                       & NMI                       & AMI                       & ARI                       & NMI                       & AMI                       & ARI                       & NMI                       & AMI                       & ARI                       & NMI                       & AMI                       & ARI                       \\
        \hline
        BERT*     & 0.10                        & 0.07                      & 0.01                      & 0.15                      & 0.10                      & 0.04                      & 0.13                      & 0.09                      & 0.03                      & 0.14                      & 0.11                      & 0.05                      & 0.10                      & 0.08                      & 0.02                      & 0.15                      & 0.12                      & 0.05                      & 0.10                      & 0.06                      & 0.02                      & 0.11                      & 0.07                      & 0.05                      \\
        SBERT*    & 0.63                        & \underline{\textit{0.62}} & \underline{\textit{0.22}} & 0.72                      & 0.70                      & 0.36                      & \underline{\textit{0.72}} & 0.70                      & 0.31                      & 0.76                      & 0.75                      & 0.53                      & 0.63                      & 0.62                      & 0.32                      & 0.70                      & 0.69                      & 0.43                      & \underline{\textit{0.70}} & \underline{\textit{0.69}} & 0.36                      & 0.63                      & 0.61                      & 0.25                      \\
        EventX    & 0.45                        & 0.16                      & 0.07                      & 0.52                      & 0.19                      & 0.07                      & 0.48                      & 0.18                      & 0.06                      & 0.51                      & 0.20                      & 0.09                      & 0.44                      & 0.15                      & 0.06                      & 0.52                      & 0.22                      & 0.11                      & 0.49                      & 0.22                      & 0.11                      & 0.39                      & 0.10                      & 0.01                      \\
        \hline
        KPGNN*    & 0.48                        & 0.46                      & 0.10                      & 0.57                      & 0.56                      & 0.18                      & 0.54                      & 0.53                      & 0.16                      & 0.55                      & 0.56                      & 0.17                      & 0.60                      & 0.60                      & 0.28                      & 0.66                      & 0.65                      & 0.43                      & 0.60                      & 0.58                      & 0.25                      & 0.52                      & 0.50                      & 0.13                      \\
        QSGNN*    & 0.52                        & 0.46                      & 0.13                      & 0.60                      & 0.58                      & 0.19                      & 0.60                      & 0.59                      & 0.20                      & 0.61                      & 0.59                      & 0.20                      & 0.59                      & 0.58                      & 0.27                      & 0.68                      & 0.67                      & 0.44                      & 0.63                      & 0.61                      & 0.27                      & 0.51                      & 0.50                      & 0.13                      \\
        HISEvent  & \textbf{0.65}               & \textbf{0.64}             & \textbf{0.42}             & \textbf{0.77}             & \textbf{0.76}             & \textbf{0.66}             & \underline{\textit{0.72}} & \underline{\textit{0.71}} & \underline{\textit{0.44}} & \textbf{0.84}             & \textbf{0.83}             & \textbf{0.80}             & \underline{\textit{0.78}} & \underline{\textit{0.78}} & \textbf{0.86}             & \textbf{0.83}             & \textbf{0.82}             & \underline{\textit{0.75}} & \textbf{0.76}             & \textbf{0.75}             & \underline{\textit{0.61}} & \underline{\textit{0.70}} & \underline{\textit{0.69}} & \underline{\textit{0.38}} \\
        \hline
        \rowcolor[HTML]{FFCE93} 
        HyperSED  & \underline{\textit{0.64}}   & \underline{\textit{0.62}} & \textbf{0.42}             & \underline{\textit{0.73}} & \underline{\textit{0.71}} & \underline{\textit{0.64}} & \textbf{0.76}             & \textbf{0.75}             & \textbf{0.77}             & \underline{\textit{0.83}} & \underline{\textit{0.81}} & \underline{\textit{0.72}} & \textbf{0.82}             & \textbf{0.81}             & \underline{\textit{0.79}} & \underline{\textit{0.79}} & \underline{\textit{0.76}} & \textbf{0.77}             & \textbf{0.76}             & \textbf{0.75}             & \textbf{0.74}             & \textbf{0.79}             & \textbf{0.78}             & \textbf{0.88}             \\
        promotion & $\downarrow$ 0.01           & $\downarrow$ 0.02         & 0.00                      & $\downarrow$ 0.04         & $\downarrow$ 0.05         & $\downarrow$ 0.02         & $\uparrow$ 0.04           & $\uparrow$ 0.04           & $\uparrow$ 0.33           & $\downarrow$ 0.01         & $\downarrow$ 0.02         & $\downarrow$ 0.08         & $\uparrow$ 0.04           & $\uparrow$ 0.03           & $\downarrow$ 0.07         & $\downarrow$ 0.04         & $\downarrow$ 0.06         & $\uparrow$ 0.02           & 0.00                      & 0.00                      & $\uparrow$ 0.13           & $\uparrow$ 0.09           & $\uparrow$ 0.09           & $\uparrow$ 0.50           \\
        \hline
        \end{tabular}
        }
        \label{tab:online_fr}
        \end{table*}

        \subsubsection{Datasets.}

            We conduct experiments on two publicly available datasets: the English Twitter dataset (68,841 messages, 503 events) \cite{mcminn2013building} and the French Twitter dataset (64,516 messages, 257 events) \cite{mazoyer2020french}. 
            Data splitting for offline and online evaluation consistent with prior studies \cite{cao2021knowledge,cao2024hierarchical}. 
            Specifically, 
                for offline evaluation, the data is divided into training, validation, and test sets with a ratio of 7:1:2; 
                for online evaluation, the data is segmented into message blocks based on time intervals: the first block encompassed messages from the first seven days (weekly), followed by daily blocks.
            Detailed statistics regarding the message blocks are shown in Table \ref{tab:efficiency_french}.

        \subsubsection{Baselines.}
        
            We compare HyperSED with two types of SED baselines. 
            (1) Non-Graph-based methods: 
                \textbf{BERT} \cite{devlin2019bert} and \textbf{SBERT} \cite{reimers2019sentence}, which are Transformer-based pre-trained language models (PLMs) known for their strong text representation capabilities; 
                \textbf{EventX} \cite{liu2020story} leverages community detection for SED. 
            (2) Graph-based methods:
                \textbf{KPGNN} \cite{cao2021knowledge} construct homogeneous graphs and utilize Graph Attention Networks (GATs) for supervised SED. 
                \textbf{QSGNN} \cite{ren2022known} employs pseudo-labels and active learning for self-supervised SED. 
                \textbf{HISEvent} \cite{cao2024hierarchical} achieves unsupervised partitioning through hierarchical minimization of two-dimensional structural entropy.
                We set the hyperparameter $n$ to 200 to mitigate occasional deadlock issues.
            All baselines, except EventX and HISEvent, require predefined event numbers.
            For implementation details, please refer to the Appendix.

        \subsubsection{Evaluation Metrics.}
        
            In line with previous SED studies \cite{cao2024hierarchical,ren2022known}, we assess the performance of all methods using three common clustering metrics: Normalized Mutual Information (\textbf{NMI}) \cite{estevez2009normalized}, Adjusted Mutual Information (\textbf{AMI}) \cite{vinh2009information}, and Adjusted Rand Index (\textbf{ARI}) \cite{vinh2009information}. 
            These metrics measure the similarity between the detected clusters and the ground truth. 

    \subsection{Online \& Offline SED Results}

        \begin{table*}[t]
        \centering
        \caption{Graph construction time, running time, and overall time of unsupervised SED methods on French Twitter.}
        \resizebox{1\linewidth}{!}{
        \begin{tabular}{cl|ccccccccccccccccc}
        \hline
        \multicolumn{2}{c|}{Blocks}                                    & Offline & $\text{M}_{1}$ & $\text{M}_{2}$ & $\text{M}_{3}$ & $\text{M}_{4}$ & $\text{M}_{5}$ & $\text{M}_{6}$ & $\text{M}_{7}$ & $\text{M}_{8}$ & $\text{M}_{9}$ & $\text{M}_{10}$ & $\text{M}_{11}$ & $\text{M}_{12}$ & $\text{M}_{13}$ & $\text{M}_{14}$ & $\text{M}_{15}$ & $\text{M}_{16}$ \\ 
        \hline
        \multicolumn{1}{c|}{\multirow{2}{*}{Statistics}} & \# Messages & 12,902  & 5,356          & 3,186          & 2,644          & 3,179          & 2,662          & 4,200          & 3,454          & 2,257          & 3,669          & 2,385           & 2,802           & 2,927           & 4,884           & 3,065           & 2,411           & 1,107           \\
        \multicolumn{1}{c|}{}                            & \# Events   & 241     & 22             & 19             & 15             & 19             & 27             & 26             & 23             & 25             & 31             & 32              & 31              & 29              & 28              & 26              & 25              & 14              \\ 
        \hline
        \multicolumn{1}{c|}{Construction}                & HISEvent    & 1,250   & 533            & 299            & 243            & 323            & 258            & 437            & 334            & 214            & 371            & 229             & 281             & 281             & 506             & 298             & 241             & 95              \\
        \multicolumn{1}{c|}{Time (s)}                    & HyperSED    & 896     & 171            & 60             & 40             & 61             & 44             & 108            & 68             & 30             & 78             & 33              & 46              & 52              & 139             & 59              & 38              & 7               \\ 
        \hline
        \multicolumn{1}{c|}{Running}                     & HISEvent    & 25,372  & 6,800          & 1,129          & 897            & 1,727          & 556            & 2,358          & 1,713          & 422            & 1,444          & 492             & 1,130           & 628             & 3,144           & 1,001           & 662             & 218             \\
        \multicolumn{1}{c|}{Time (s)}                    & HyperSED    & 65      & 25             & 58             & 11             & 12             & 17             & 17             & 16             & 17             & 21             & 16              & 24              & 17              & 25              & 64              & 11              & 6               \\ 
        \hline
        \multicolumn{1}{c|}{Overall}                     & HISEvent    & 26,622  & 7,333          & 1,428          & 1,140          & 2,050          & 814            & 2,795          & 2,047          & 636            & 1,815          & 721             & 1,411           & 909             & 3,650           & 1,299           & 903             & 313             \\
        \multicolumn{1}{c|}{Time (s)}                    & HyperSED    & 961     & 196            & 118            & 51             & 73             & 61             & 125            & 84             & 47             & 99             & 49              & 70              & 69              & 164             & 123             & 49              & 13              \\ 
        \hline
        \multicolumn{2}{c|}{Speed-up (times $\times$)}                 & 27.70   & 37.41          & 12.10          & 22.35          & 28.08          & 13.34          & 22.36          & 24.37          & 13.53          & 18.33          & 14.71           & 20.16           & 13.17           & 22.26           & 10.56           & 18.43           & 24.08           \\ 
        \hline
        \end{tabular}
        }
        \label{tab:efficiency_french}
        \end{table*}
        
        \textbf{Offline:}
        The results for the offline SED are presented in Table \ref{tab:offline}, demonstrating that the proposed method, HyperSED, achieves superior performance on both datasets, surpassing existing methods by an average of 0.02, 0.04, and 0.27 in NMI, AMI, and ARI, respectively.
        The relatively poor performance of HISEvent compared to its online results can be attributed to the larger scale of messages and more diverse events in the offline setting. 
        These factors impede HISEvent from hierarchically determining the minimum structural entropy.
        Our approach, on the other hand, is theoretically immune to the influence of data scale and demonstrates great performance in handling diverse events, as indicated by the higher ARI score in comparison to other baselines.
        
        \noindent\textbf{Online:} 
        The online results for English and French Twitter are shown in Table \ref{tab:online_en} and Table \ref{tab:online_fr}. 
        The results reveal that HyperSED demonstrates competitive performance across the majority of message blocks, with a marginal decrease in some blocks.
        Notably, within English Twitter, significant improvements are observed in message blocks $\text{M}_{1}$, $\text{M}_{7}$, and $\text{M}_{15}$. 
        The statistical results show that these blocks contain larger message scales and more diverse events. 
        The observed marginal decrease is likely due to error accumulation during SAMG construction, where semantically similar messages are incorrectly clustered under a single anchor. 
        This decrease represents a trade-off between effectiveness and efficiency, with the improvements considered worthwhile.

        \begin{figure*}[!t]
            \centering
            \includegraphics[width=1.0\linewidth]{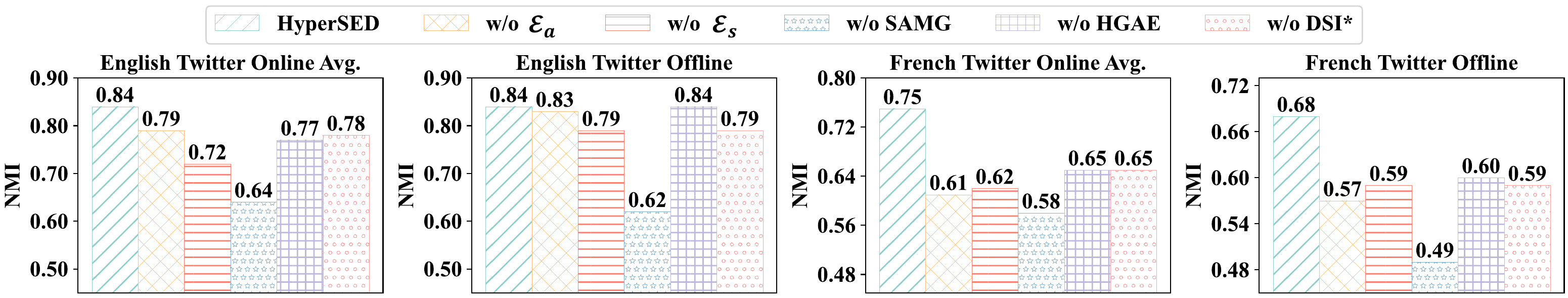}
            \caption{Ablation results for English and French Twitter. Online results are the average results across all blocks.}
            \label{fig:ablation}
        \end{figure*}
        
        \noindent\textbf{Key Insights:} 
        It is observed that HyperSED excels when confronted with larger message sizes and diverse events. 
        We attribute this success to the proposed SAMG, where core semantics and interrelations are preserved.

    \subsection{Efficiency Analysis of Unsupervised SED}\label{sec:exp_eff}
        
        Table \ref{tab:efficiency_french} presents the graph construction time, running time, and total time of the unsupervised SED baseline HISEvent and the proposed HyperSED on the French Twitter dataset.
        
        \noindent\textbf{HyperSED:}
        HyperSED exhibits remarkably short running times, even with the largest amount of messages (offline setting), which takes up only 65 seconds, whereas HISEvent takes 25,372 seconds.
        Compared to its running time, its graph construction time is longer, especially in the offline setting; however, consider the data scale, is all reasonable.
        
        \noindent\textbf{Comparison:}
        In contrast, both the graph construction time and running time of HISEvent significantly lag behind HyperSED across all blocks.
        Overall, HyperSED demonstrates an average speedup (online setting) of 21.04 times in total time over HISEvent, with a maximum improvement of 37.41 times.
        Efficiency is paramount for SED systems as they must promptly respond to new messages and events. 
        In this context, the proposed HyperSED outperforms existing unsupervised SED methods to a significant extent.

    \subsection{Ablation Study}

        The results of the ablation study are presented in Figure \ref{fig:ablation}. 
        With the absence of DSI, the method requires the total event number as the supervision signal.
        
        \noindent\textbf{Maximum Influence:}
        The largest performance drop is observed in the absence of SAMG.
        This is because, in scenarios of large-scale messages, the model without SAMG needs to learn all messages and their relationships, which is susceptible to noise interference. 
        SAMG not only reduces the node size but also simplifies their relationships, which mitigates the influences caused by noise to a great extent.
        
        \noindent\textbf{Minimum Influences:}
        The least performance drop is observed when any graph edge $\mathcal{E}_{a}$ or $\mathcal{E}_s$ is absent.
        This is due to SAMG solely preserving edges between messages not belonging to the same anchor.
        In most cases, messages sharing high similarity or attribute co-occurrence are housed within the same anchor, thereby limiting its impact.
        Nonetheless, the construction of these edges remains indispensable to the overall performance of HyperSED.

    \subsection{Parameter Sensitivity Analysis}
    
        \begin{figure}[!ht]
            \centering
            \includegraphics[width=1.0\linewidth]{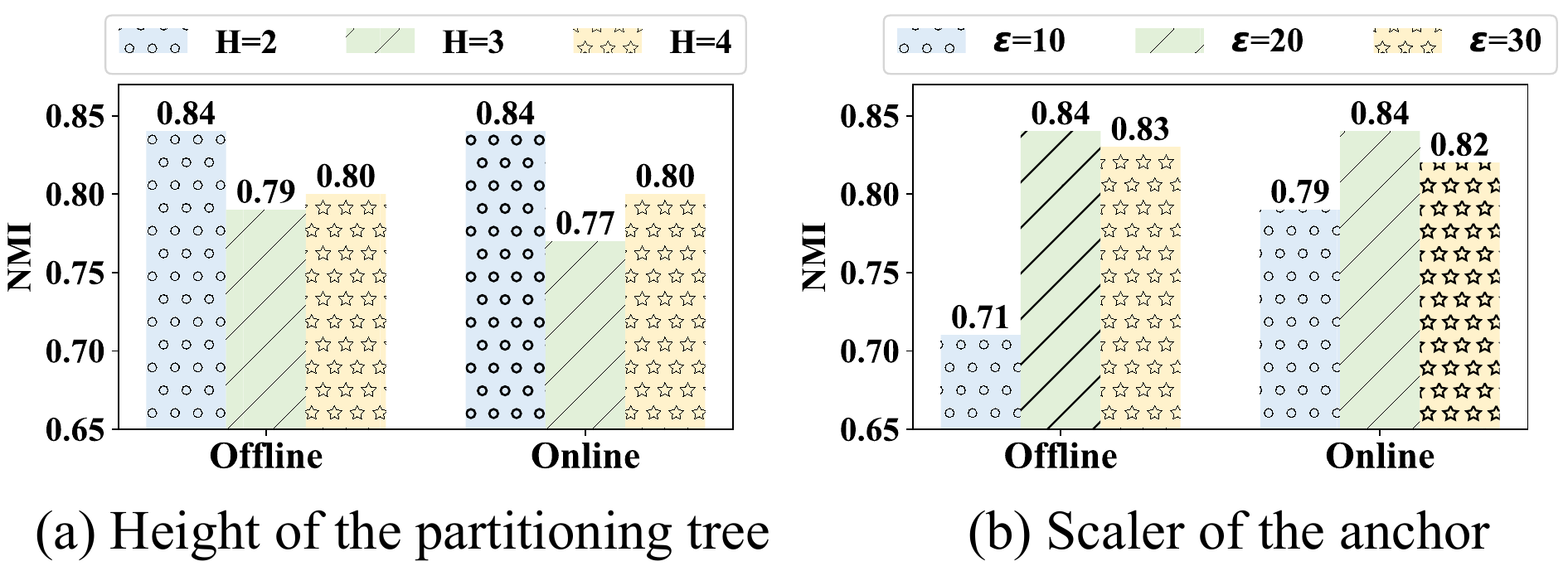}
            \caption{Results of parameter sensitivity analysis for English Twitter. Online results are the average results across all blocks.}
            \label{fig:para_sen}
        \end{figure}
        
        We analyze the parameter sensitivity concerning the dimension of structural entropy (height of the partitioning tree) $H=\{2,3,4\}$ and the number of anchor nodes $M=N/\epsilon, \epsilon=\{10,20,30\}$.
        The results on the English Twitter dataset are shown in Figure \ref{fig:para_sen}.
        The different parameters show minor fluctuations, all of which demonstrate reasonable performance without significant deviations.


\section{Conclusion}

    This work proposes HyperSED, an unsupervised SED framework that models social messages as anchors and learns their interrelations within the hyperbolic space.
    HyperSED learned semantically representative anchor nodes and simplified the relations among messages by incorporating SAMG.
    A substantial reduction in the detection runtime is achieved through the collaboration of SAMG and learning in the hyperbolic space.
    Experiments on public datasets demonstrate the effectiveness and efficiency of HyperSED, achieving competitive performance and enhancing efficiency significantly.


\section{Acknowledgments}
We are sincerely grateful to all the reviewers and chairs for dedicating their valuable time and providing insightful suggestions.
This work is supported by the Yunnan Provincial Major Science and Technology Special Plan Projects (No.202302AD080003).
Hao Peng is supported by NSFC through grant 62322202.


\bibliography{aaai25}

\clearpage
\appendix

\section{Structural Entropy}

    \begin{definition}[H-dimensional Structural Entropy \cite{li2016structural}]
        Given a weighted graph $G=(\mathcal{V}, \mathcal{E})$ with weight function $w$ and a partitioning tree $\mathcal{T}$ of $G$ with height $H$,
         the \emph{structural information} of $G$ with respect to each non-root node $\alpha$ of $\mathcal{T}$ is defined as
        \begin{align}
        \label{eq.si}
            \mathcal{H}^{\mathcal{T}}(G;\alpha)=-\frac{g_\alpha}{\operatorname{vol}(G)}\log_2\frac{\operatorname{vol}(T_\alpha)}{\operatorname{vol}(T_{\alpha^-})}.
        \end{align}
        In the \emph{partitioning tree} $\mathcal{T}$ with root node $\lambda$, each tree node $\alpha$ is associated with a subset of $\mathcal V$, denoted as module $T_\alpha$, and the immediate predecessor of it is written as $\alpha^-$. The module of the leaf node is a singleton of the graph node. 
        The scalar $g_\alpha$ is the total weights of graph edges with exactly one endpoint in module $T_\alpha$.
        Then, the $H$-dimensional structural information of $G$ by $\mathcal{T}$ is given as,
        \begin{align}
        \label{si}
            \mathcal{H}^{\mathcal{T}}(G) = \sum\nolimits_{\alpha \in \mathcal{T}, \alpha \neq \lambda }\mathcal{H}^{\mathcal{T}}(G;\alpha).
        \end{align}
        Traversing all possible partitioning trees of $G$ with height $H$, \emph{$H$-dimensional structural entropy} of $G$ is defined as
        \begin{align}
            \mathcal{H}^{H}(G) = \min_\mathcal{T} \mathcal{H}^{\mathcal{T}}(G), \quad
        \label{eq.opt_tree}
            \mathcal{T}^*=\underset{\mathcal{T}}{\arg\min} \mathcal{H}^{\mathcal{T}}(G),
        \end{align}
        where  $\mathcal{T}^*$ is the optimal tree of $G$ which encodes the self-organization and minimizes the uncertainty of the graph.
    \end{definition}

\section{Hyperbolic Space}

    \subsection{Riemannian Manifold}
    
        Riemannian manifold is a real and smooth manifold $\mathbb{M}$ equipped with Riemannian metric tensor $g_{\boldsymbol x}$ on the tangent space $\mathcal{T}_{\boldsymbol{x}}\mathbb{M}$ at point $\boldsymbol{x}$.
        A Riemannian metric assigns to each $\boldsymbol{x} \in \mathbb{M}$ a positive-definite inner product $g_{\boldsymbol{x}}: \mathcal{T}_{\boldsymbol{x}}\mathbb{M} \times \mathcal{T}_{\boldsymbol{x}}\mathbb{M} \rightarrow \mathbb{R}$, which induces a norm defined as $\lvert \cdot \rvert: \boldsymbol{v} \in \mathcal{T}_{\boldsymbol{x}}\mathbb{M} \mapsto \sqrt{g_{\boldsymbol{x}}(\boldsymbol{v}, \boldsymbol{v})} \in \mathbb{R}$.
        
        An exponential map at point $\boldsymbol x \in \mathbb{M}$ is denoted as $\exp_{\boldsymbol x}(\cdot): \boldsymbol u \in \mathcal{T}_{\boldsymbol{x}}\mathbb{M} \mapsto \exp_{\boldsymbol x}(\boldsymbol u) \in \mathbb{M}$. It takes a tangent vector $\boldsymbol u$ in the tangent space at $\boldsymbol x$ and transforms $\boldsymbol x$ along the geodesic starting at $\boldsymbol x$ in the direction of $\boldsymbol u$.
        
        A logarithmic map at point $\boldsymbol x$ is the inverse of the exponential map $\boldsymbol x$, which maps the point $\boldsymbol y \in \mathbb{M}$ to a vector $\boldsymbol{v}$ in the tangent space at $\boldsymbol x$.

        \begin{table*}[!t]
        \centering
        \caption{Graph construction time, running time, and overall time of unsupervised SED methods on English Twitter.}
        \resizebox{1\linewidth}{!}{
        \begin{tabular}{cl|ccccccccccc}
        \hline
        \multicolumn{2}{c|}{Blocks}                                    & Offline         & $\text{M}_{1}$  & $\text{M}_{2}$  & $\text{M}_{3}$  & $\text{M}_{4}$  & $\text{M}_{5}$  & $\text{M}_{6}$  & $\text{M}_{7}$  & $\text{M}_{8}$  & $\text{M}_{9}$  & $\text{M}_{10}$ \\
        \hline
        \multicolumn{1}{c|}{\multirow{2}{*}{Statistics}} & \# Messages & 13769           & 8722            & 1.491           & 1835            & 2010            & 1834            & 1276            & 5278            & 1560            & 1363            & 1096            \\
        \multicolumn{1}{c|}{}                            & \# Events   & 488             & 41              & 30              & 33              & 38              & 30              & 44              & 57              & 53              & 38              & 33              \\ 
        \hline
        \multicolumn{1}{c|}{Construction}                & HISEvent    & 1290            & 836             & 126             & 156             & 175             & 158             & 104             & 521             & 132             & 109             & 91              \\
        \multicolumn{1}{c|}{Time (s)}                    & HyperSED    & 1030            & 504             & 19              & 25              & 28              & 24              & 11              & 177             & 17              & 12              & 9               \\ 
        \hline
        \multicolumn{1}{c|}{Running}                     & HISEvent    & 14649           & 19471           & 471             & 327             & 379             & 254             & 234             & 5779            & 388             & 358             & 242             \\
        \multicolumn{1}{c|}{Time (s)}                    & HyperSED    & 122             & 26              & 8               & 8               & 14              & 8               & 8               & 23              & 8               & 7               & 7               \\ 
        \hline
        \multicolumn{1}{c|}{Overall}                     & HISEvent    & 15939           & 20307           & 597             & 483             & 554             & 412             & 338             & 6300            & 520             & 467             & 333             \\
        \multicolumn{1}{c|}{Time (s)}                    & HyperSED    & 1152            & 530             & 27              & 33              & 42              & 32              & 19              & 200             & 25              & 19              & 16              \\ 
        \hline
        \multicolumn{2}{c|}{Speed-up (times $\times$)}                 & 13.84           & 38.32           & 22.11           & 14.64           & 13.19           & 12.88           & 17.79           & 31.50           & 20.80           & 24.58           & 20.81           \\ 
        \hline
        \multicolumn{2}{c|}{Blocks}                                    & $\text{M}_{11}$ & $\text{M}_{12}$ & $\text{M}_{13}$ & $\text{M}_{14}$ & $\text{M}_{15}$ & $\text{M}_{16}$ & $\text{M}_{17}$ & $\text{M}_{18}$ & $\text{M}_{19}$ & $\text{M}_{20}$ & $\text{M}_{21}$ \\
        \hline
        \multicolumn{1}{c|}{\multirow{2}{*}{Statistics}} & \# Messages & 1232            & 3237            & 1972            & 2956            & 2549            & 910             & 2676            & 1887            & 1399            & 893             & 2410            \\
        \multicolumn{1}{c|}{}                            & \# Events   & 30              & 42              & 40              & 43              & 42              & 27              & 35              & 32              & 28              & 34              & 32              \\ 
        \hline
        \multicolumn{1}{c|}{Construction}                & HISEvent    & 105             & 290             & 183             & 271             & 252             & 70              & 249             & 165             & 115             & 69              & 214             \\
        \multicolumn{1}{c|}{Time (s)}                    & HyperSED    & 12              & 66              & 36              & 63              & 61              & 7               & 61              & 26              & 14              & 5               & 36              \\ 
        \hline
        \multicolumn{1}{c|}{Running}                     & HISEvent    & 220             & 620             & 793             & 703             & 357             & 109             & 684             & 802             & 187             & 101             & 1471            \\
        \multicolumn{1}{c|}{Time (s)}                    & HyperSED    & 9               & 12              & 15              & 11              & 17              & 4               & 11              & 8               & 5               & 5               & 12              \\ 
        \hline
        \multicolumn{1}{c|}{Overall}                     & HISEvent    & 325             & 910             & 976             & 974             & 609             & 179             & 933             & 967             & 302             & 170             & 1685            \\
        \multicolumn{1}{c|}{Time (s)}                    & HyperSED    & 21              & 78              & 51              & 74              & 78              & 11              & 72              & 34              & 19              & 10              & 48              \\ 
        \hline
        \multicolumn{2}{c|}{Speed-up (times $\times$)}                 & 15.48           & 11.67           & 19.14           & 13.16           & 7.81            & 16.27           & 12.96           & 28.44           & 15.89           & 17.00           & 35.10           \\
        \hline
        \end{tabular}
        }
        \label{tab:efficiency_en}
        \end{table*}
        
        \subsection{Poincar\'{e} ball model}

            The Riemannian metric tensor at $\boldsymbol{x}$ is $g_{\boldsymbol x}^{\mathbb{B}_\kappa}=(\lambda_{\boldsymbol{x}}^\kappa)^2g^\mathbb{E}_{\boldsymbol{x}}$, where $\lambda_{\boldsymbol{x}}^\kappa=\frac{2}{1+\kappa\lVert \boldsymbol{x} \rVert^2}$ and $g^\mathbb{E}_{\boldsymbol{x}}=\mathbf{I}$ is the Euclidean metric.

\section{Technical Details}

    \subsection{Implementation Details}
    
        The experiments are implemented using the PyTorch framework and run on a machine with eight NVIDIA Tesla A100 (40G) GPUs.
        The baseline methods are implemented with their original configurations.
        For HISEvent, we set the hyperparameter $n$ to 200 to mitigate occasional deadlock issues and reduce computational runtime.
        For SBERT, we adopt `all-MiniLM-L6-v2' to obtain the initial message representation for English Twitter (384 dimensions) and `distiluse-base-multilingual-cased-v1' for French Twitter (512 dimensions).
        We set the training epochs to 200, with an early stop patience of 50, the learning rate to $1e-3$, and the dropout rate to 0.4.
        For message construction, the threshold search space is set to $\Pi=[0.4, 0.6]$ and the search step to 0.05.
        For SAMG, the anchor number is determined based on the number of messages as $M=N/\epsilon$, where $\epsilon=\{20, 30, 40\}$.
        For the HGAE, the number of layers is set to 2, the hidden feature size is set to 128, and the output feature size is set to 64.
        For DSI, the number of layers is set to 2, the hidden feature size to 64, the output feature size to 3, the height of the encoding tree to 2, and $N_{h-1}=500$.
        For BERT, we adopt `base-multilingual-cased' for both languages (768 dimensions).
        For BERT and SBERT, PLMs are utilized to obtain message representations, followed by k-means clustering to derive results. 
        Baselines requiring a predefined number of events for clustering are configured with the ground truth number (a necessity for evaluation but impractical in real-world scenarios), marked in the result tables with `*'.
        We report the average score over 5 individual runs.

    \subsection{Efficiency Analysis on the English Twitter}

        The results for graph construction time, running time, and overall time on the English Twitter Dataset are presented in Figure \ref{tab:efficiency_en}.
        Consistent with the results on the French Twitter dataset, HyperSED's progress demonstrates a significant lead in efficiency compared to the unsupervised SED baseline HISEvent. 
        Specifically, in message block $\text{M}{1}$, HyperSED surpasses HISEvent by a substantial margin of 38.32 times and, at a minimum, outperforms HISEvent by 7.81 times in message block $\text{M}{15}$.
        Even in the offline scenario with the highest message size, the running time is 1152 seconds, a 120 times improvement over HISEvent's runtime.

    \subsection{Case Study \& Visualization}

\begin{figure*}[!t]
            \centering
            \includegraphics[width=1.0\linewidth]{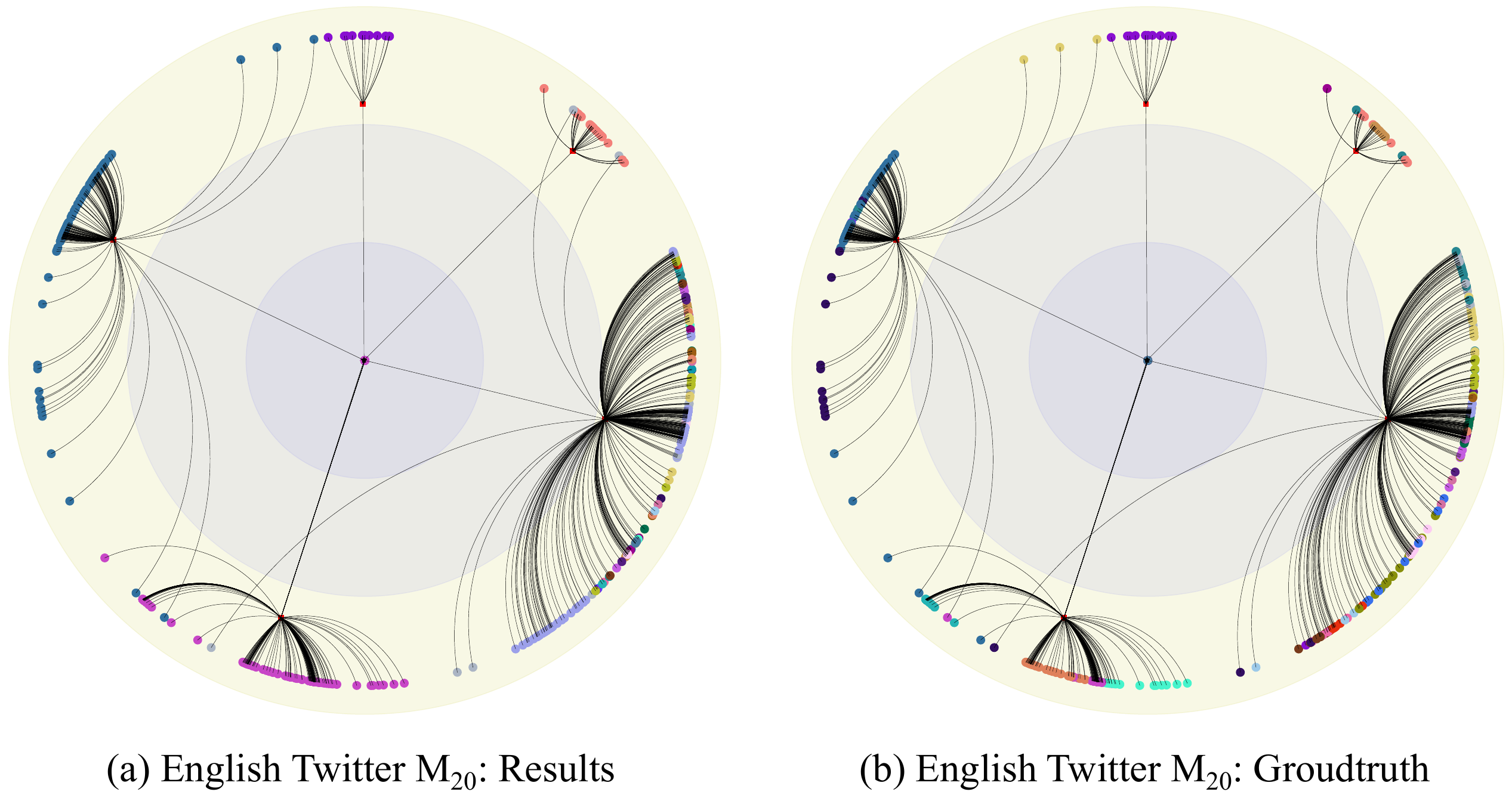}
            \caption{Caption}
            \label{fig:visual}
        \end{figure*}

        We conduct a case study on the message block $\text{M}_{20}$ from the English Twitter dataset, utilizing the 'w/o SAMG' to construct the hyperbolic partitioning tree and visualize the results, as it provides a message-wise view.
        We apply stereographic projection $\Psi$ \cite{bachmann2020constant} to the Poincar'e ball model to obtain the corresponding Poincar\'{e} disc, which is preferable to visualization.
        The visualization figures are shown in Figure \ref{fig:visual}, Figure \ref{fig:visual}(a) is the result of the message-wise partitioning tree, and Figure \ref{fig:visual}(b) is the ground-truth of the message nodes.
        The findings indicate that the learned clusters exhibit clear separation and closely match the ground-truth clusters.
        In the constructed partitioning tree, the identified event illustrated in Figure \ref{fig:visual}(a) appears to show fewer distinctions compared to those in the ground truth, as evidenced by the prevalence of dark pink nodes at the 7 o'clock position. 
        Conversely, Figure \ref{fig:visual}(b) exhibits a range of colors, implying increased diversity.
        This disparity may arise from the similarity in their representations, where the close proximity of nodes to each other is visually discernible in the disc representation.

\end{document}